\documentclass[10pt,twocolumn,letterpaper]{article}

\usepackage{iccv}
\usepackage{times}
\usepackage{epsfig}
\usepackage{graphicx}
\usepackage{amsmath}
\usepackage{amssymb}
\usepackage{colortbl}

\usepackage{xcolor}
\usepackage{enumitem}
\setitemize{noitemsep,topsep=0pt,parsep=0pt,partopsep=0pt}
\usepackage{soul}
\usepackage{booktabs}

\usepackage[breaklinks=true,bookmarks=false]{hyperref}

\iccvfinalcopy 

\newcommand{\comment}[1]{}

\newcommand{\bD}{\mathbf{D}}

\newcommand{\bM}{\mathbf{M}}

\newcommand{\bS}{\mathbf{S}}

\newcommand{\mB}{\mathcal{B}}

\newcommand{\mE}{\mathcal{E}}

\newcommand{\mL}{\mathcal{L}}

\ificcvfinal\fi
\begin{document}

\title{Human Motion Prediction via Spatio-Temporal Inpainting}

\author{A. Hernandez Ruiz$^{1}$ \hspace{0.7cm}
J.  Gall$^{2}$ \hspace{0.7cm}
F.  Moreno-Noguer$^{1}$ \\
$^{1}$Institut de Rob\`otica i Inform\`atica Industrial, CSIC-UPC, Barcelona, Spain\\
$^{2}$ Computer Vision Group, University of Bonn, Germany \\
}

\maketitle

\begin{abstract}
We propose a Generative Adversarial Network (GAN) to forecast 3D human motion given a sequence of past  3D skeleton poses. While recent  GANs have shown  promising results, they can only  forecast plausible  motion over relatively short periods of time (few hundred milliseconds) and typically ignore the absolute position of the skeleton w.r.t. the camera. Our scheme  provides long term predictions (two seconds or more) for both the  body pose and its absolute position. Our approach builds upon three main contributions. First, we represent the data using a spatio-temporal tensor of 3D skeleton coordinates which allows formulating the prediction problem as an inpainting one, for which GANs work particularly well. Secondly, we  design an architecture   to learn the joint distribution of body poses and global motion,  capable to hypothesize large chunks of the input 3D tensor with missing data. And finally, we argue that the L2 metric,  considered so far by most approaches, fails to capture the actual distribution of long-term human motion. We propose two alternative metrics, based on the distribution of frequencies, that are able to capture more realistic motion patterns.   Extensive experiments demonstrate  our approach to significantly improve   the state of the art, while also handling situations in which past observations are corrupted by  occlusions, noise and  missing frames. 
\end{abstract}

\section{Introduction}
\label{sec:introduction}

\begin{figure}[t!]
    \begin{tabular}{cc}
    \hspace{0.7cm}{\bf Ground Truth} & \hspace{1.9cm}{\bf Generated}
    \end{tabular}
    \includegraphics[width=0.47\textwidth]{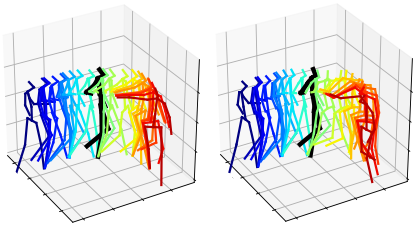}
    \caption{{\bf Example result.} Our approach is the first in generating full body pose, including skeleton motion and absolute position in space. The predicted sequence starts from the skeleton marked in black. Note that the generated motion is somewhat different but semantically indistinguishable from the ground truth.}
    \label{fig:teaser}
\end{figure}

Recent advances in motion capture  technologies, combined with large scale datasets such as Human3.6M~\cite{h36m_pami}, have spurred the interest for new deep learning algorithms able to forecast 3D human motion from past skeleton data. State-of-the-art approaches formulate the problem as a sequence generation task, and solve it using Recurrent Neural Networks (RNNs)~\cite{fragkiadaki2015recurrent, jain2016structural}, sequence-to-sequence models~\cite{martinez2017human} or encoder-decoder predictors~\cite{butepage2018anticipating, gui2018adversarial}. While promising results, these works suffer from three fundamental limitations. First, they address a simplified version of the problem in which global body positioning is disregarded, either by parameterizing 3D body joints using position agnostic angles~\cite{fragkiadaki2015recurrent, jain2016structural, martinez2017human} or  body centered  coordinates~\cite{butepage2018anticipating}. Second, current methods require additional supervision in terms of action labels  during training and inference, which limits their generalization capabilities. And third, most approaches aim to minimize the L2 distance between the ground truth and generated motions. The L2 distance, however, is known to be an inaccurate metric, specially to compare long motion sequences. In particular, the use of this metric to train a deep network favors motion predictions that converge to a static mean pose. Even though this issue has been raised in~\cite{gui2018adversarial,jain2016structural} and is partially solved during training using other metrics (e.g. geodesic loss), the L2 distance is still being used as a common practice when benchmarking different methodologies. To our understanding, this practice compromises the progress in this field.

In this paper we tackle all three issues. Specifically, we design a novel GAN architecture that is conditioned on past observations and is able to jointly forecast non-rigid body pose and its absolute position in space. For this purpose we represent the observed skeleton poses (expressed in the camera reference frame) using a spatio-temporal tensor and formulate the prediction problem as an inpainting task, in which a part of the spatio-temporal volume needs to be regressed. A GAN architecture consisting of a fully convolutional generator specially designed  to preserve the temporal coherence, and three independent discriminators that enforce anthropomorphism of the generated skeleton and its motion, makes it possible to render highly realistic long-term predictions (of 2 seconds or more). Interestingly, the L2 loss is only enforced over the reconstructed past observations and not over the hypothesized future predictions.  This way, the generation of future frames is fully controlled by the discriminators. In fact, our model does not require  ground truth annotations of the generated frames nor explicit information about the action being performed.

We also introduce a novel metric for estimating the similarity between the generated and the ground truth sequences. Instead of seeking to get a perfect match for all joints across all frames (as done when using the L2 distance), the metric we propose aims to estimate the similarity between distributions over the human motion manifold.

In the experimental section we show that our approach, besides yielding full body pose, orientation and position, is also robust to challenging artifacts including missing frames and occluded joints on the past skeleton observations. Fig.~\ref{fig:teaser} shows an example result of our approach.

\section{Related Work}
\label{sec:related}

%

\vspace{1mm}
\noindent{\bf Deep Learning for motion prediction.} Most recent deep learning approaches build upon the problem formulation proposed by~\cite{taylor2007modeling} in which input motion sequences are represented by 3D body joint angles in a kinematic tree. Motivated by their success in machine translation problems~\cite{cho2014learning,karpathy2015visualizing,sutskever2014sequence}, RNNs are then used to predict motion sequences of body joint angles. For instance, Fragkiadaki \etal~\cite{fragkiadaki2015recurrent} introduce an Encoder-Recurrent-Decoder (ERD) in combination with a Long Short-Term Memory (LSTM) for this purpose. Jain \etal~\cite{jain2016structural} introduced structural RNNs, an approach that exploits the structural hierarchy of the human body parts. Martinez \etal~\cite{martinez2017human} develop a sequence-to-sequence architecture with a residual connection that incorporates action class information via one-hot vectors. While well-suited for the particular motion they are trained for, these approaches do not generalize to other actions. More importantly, these models are only effective at predicting in the short- and mid-range time horizons, and are often surpassed by a simple zero velocity baseline model. This is in part due to the use of the L2 metric both for training and evaluation. More recent approaches have different strategies to address this problem.

Li \etal~\cite{li2018convolutional}, propose a model with an auto-regressive CNN generator, and combine the L2 loss with an adversarial loss.
Gui \etal~\cite{gui2018adversarial} fully eliminate the L2 loss at training, and utilize an ERD model with a combination of an adversarial and geodesic losses. These works, however, still perform evaluation according to the L2 metric, which fails to capture the semantics of the motion, specially for long term predictions. Furthermore, since motion is parameterized in terms of the joint angles, the rotation and translation of the body in space is not estimated. 


\vspace{1mm}
\noindent\textbf{Sequence completion and image inpainting}. 
Completing missing data within a sequence has been traditionally addressed using low-rank matrix factorization~\cite{agudo2017dust,xia2018nonlinear}. Deep Learning approaches have also been used for this purpose \eg, through RNNs~\cite{kucherenko2018neural,mall2017deep}. These works, however, are not designed for future prediction. 

Image inpainting is a very related problem. In the deep learning era, Denoising AEs~\cite{bengio2013generalized} and Variational AEs~\cite{kingma2013auto} have become prevalent frameworks for denoising and completing missing data and image inpainting. These baselines, however, cannot handle large missing portions of structured data.
State of the art has been significantly pushed by GANs conditioned with partial or corrupted images ~\cite{radford2015unsupervised,yeh2017semantic,yu2018generative}. As we shall see, our approach draws inspiration on this idea.

\begin{figure*}[t!]
    \centering
    \includegraphics[width=0.9\textwidth]{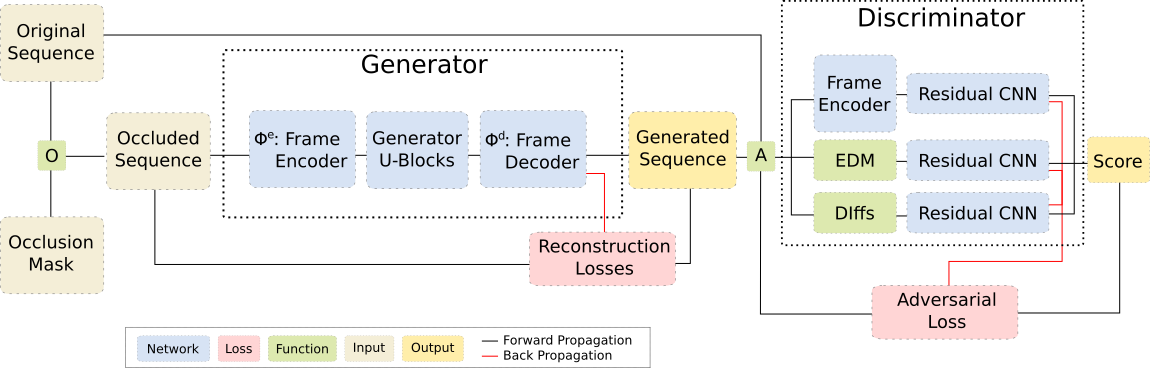}
    \caption{Overview of our 
    architecture. An input masked sequence of 3D joint coordinates is fed into a fully convolutional and time preserving generator. The output sequence is controlled by  a number of geometric constraints, including losses applied to the generator output and adversarial losses of three independent discriminators. }
    \label{fig:motiongan}
    \vspace{-2mm}
\end{figure*}

\vspace{1mm}
\noindent\textbf{Metrics for evaluating human motion prediction.} 
The fact that L2 is not appropriate for measuring similarity between human motion sequences has been recently discussed and addressed in different works. Coskun \etal~\cite{coskun2018human} use deep metric learning and a contrastive loss to learn the metric directly from the data. This is arguably the best way to compare the motion sequences semantically. Nonetheless, the problem with this approach is that once the metric is trained, it is hard to apply to different models, since the metric is trained with a specific setup. An alternative for sequences that does not require training, would be to use frequency based metrics. In~\cite{gopalakrishnan2018neural} a metric based in power spectrum is proposed. This metric shows interesting properties, and seems suitable to compare actions with periodic motion such as walking. The main drawback of this approach is that it compares sequence by sequence, which in our view is undesirable.  We, instead, would like to compare distributions of sequences.


For image generation, there are recent works that propose measuring the fitness of the models based on properties of the distribution of the generated data. The Inception Score~\cite{salimans2016improved} measures the entropy of the label outputs of the inception network on the generated images. The Frechet Inception Distance~\cite{heusel2017gans} (FID) propose instead to fit two multivariate Gaussians to the activations of the inception network, when the real and generated samples respectively. Then the FID is obtained by measuring the distance between the gaussian models. 
Following these works, we propose new metrics based on the distribution of the frequencies of the generated samples. These metrics have the advantage of being easy to implement and replicate, and they measure the general fitness of a model by taking into account a distribution of sequences.

\vspace{-1mm}
\section{Problem Formulation}
\label{sec:problem}
We represent human pose with a $J$-joint skeleton, where each joint consists of its 3D Cartesian coordinates expressed in the camera  reference frame. Rotation and translation transformations are inherently encoded in such coordinates. A motion sequence is  a concatenation of $F$ skeletons, which we shall represent by a tensor $\bS \in {\mathbb R}^{\mathrm{F\times{J}\times{3}}}$. Let us define an occlusion mask as a binary matrix  $\bM \in {\mathbb B}^{\mathrm{F\times{J}\times{3}}}$, which determines the part of the sequence that is not observed and is applied onto the sequence by performing the element-wise dot product $\bS \circ \bM \equiv \bS^\mathrm{m}$. Our goal is then to estimate the 3D coordinates of the masked joints.
Note that depending on the pattern used to generate the occlusion mask $\bM$, we can define different sub-problems. For instance, by masking the last frames of the sequence we can represent a forecasting problem; if instead we mask specific intermediate joints, then we represent random joints occlusions, structured occlusions or missing frames. Our model can tackle any combination of these sub-problems. 

\vspace{-1mm}
\section{Model}
\label{sec:model}
\vspace{-1mm}
\subsection{STMI-GAN architecture}
Fig.~\ref{fig:motiongan} shows an overview of the  GAN we propose, in which we pose the human motion prediction problem as an inpainting task in the spatio-temporal domain. We denote our network as STMI-GAN (Spatio-Temporal Motion Inpainting GAN). We next describe its main components.

\vspace{1mm}
\noindent{\bf Generator}.
The rationale for the design holds in that convolutional GANs have been successful in image inpainting problems, which is similar to  ours. A masked human motion sequence $\bS^\mathrm{m}$, however, cannot be directly processed by a convolutional network because the dimension corresponding to the joints ($J$) does not have a spatial meaning, in contrast to the temporal ($F$) and Cartesian coordinates  dimensions. That is, neighboring joints along this dimension do not correspond to neighboring joints in 3D space\footnote{For instance, the joint \#0 is normally the hip, and its neighbors in the body graph are the joints \#1 (left hip), \#5 (right hip) and \#9 (spine).}. 
To alleviate this  lack of spatial continuity problem, the generator is placed in-between a frame autoencoder, namely a frame encoder $\Phi^e$ and a frame decoder $\Phi^d$, which are symmetrical networks. The frame encoder,  operates over the J-dimension and projects each frame $\bS^\mathrm{m}_{i::}\in {\mathbb R}^{\mathrm{J\times{3}}} $ of the sequence to a one-dimensional vector $\bS^\mathrm{e}_{i::}=\Phi^e(\bS^\mathrm{m}_{i::})\in {\mathbb R}^{\mathrm{H\times{1}}}$, where $H$ is the dimension of the space for the pose embedding\footnote{ $\bS_{i::}$ denotes the $i$-th element of $\bS$ along the first axis.}. To project each frame, the encoder does not use information from neighboring frames, being thus   time invariant. 
We denote the encoded sequence as $\bS^e \in {\mathbb R}^{\mathrm{F\times{H}\times{1}}}$. 

\begin{figure}[h!]
\begin{centering}
    \includegraphics[width=0.45\textwidth]{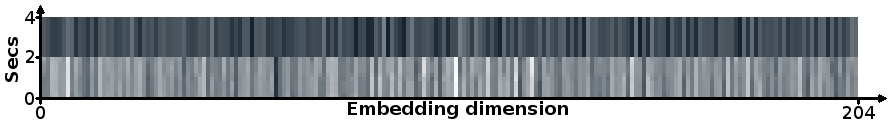}
    \caption{{\bf Motion embedding.} A motion sequence of the H3.6M dataset passed through the frame encoder. The sequence is  occluded from the half onwards, with the goal of motion prediction. }
    \label{fig:embedding}
\end{centering}
\end{figure}

As we can observe in Fig.\ref{fig:embedding}, the frame encoder learns to represent the sequence as a 2D matrix, in an arbitrary space. Although the learnt space has no clear interpretation, we can observe from the sample that it retains certain properties, such as the temporal ordering, and a constant "zero" value for the occluded frames. This encoded sequence is then passed through a series of generator blocks $\Phi^g$, which produces a new sequence $\bS^g \in {\mathbb R}^{\mathrm{F\times{H}\times{1}}}$ in the embedded space. The blocks of the generators are CNNs that process the sequence in both temporal and spatial dimensions. Further details are explained in Section \ref{sec:implementation}. Finally the decoder network $\Phi^d$ maps back the the transformed sequence $\bS^g$ to the output sequence in the original shape $\bS^{out} \in {\mathbb R}^{\mathrm{F\times{J}\times{3}}}$.

\vspace{1mm}
\noindent{\bf Discriminator}.
To capture the complexity of the human motion distribution we  split the discriminator into three branches, capturing  different aspects of the generated sequence. Each discriminator is a Residual CNN classifier that serves as a feature extractor. These features are linearly combined to obtain a probability of the sequence of being real. We next describe the main blocks of our model. Details of the underlying architectures are detailed later in Sect.~\ref{sec:implementation}.

\vspace{1mm}
\noindent{\bf Base discriminator}. The same architecture of the frame encoder $\Phi^e$, with independent parameters, is used to process the generated sequence $\bS^{out}$. The reason to reuse such architecture is that we want a discriminator to be applied directly on the non Euclidean representation   used by the CNN blocks of the generator,  in order to boost its performance.

\vspace{1mm}
\noindent{\bf EDM discriminator}. We introduce a geometric discriminator that evaluates the anthropomorphism of the generated sequence $\bS^{out}$ via the analysis of its Euclidean Distance Matrix, computed as   $\textrm{EDM}(\bS^{out})\equiv \bD\in {\mathbb R}^{\mathrm{J\times{J}}}$, where $\bD_{ij}$ is the Euclidean distance between joints $i$ and $j$ of $\bS^{out}$. This is  a rotation and translation invariant representation~\cite{hernandez20173d,moreno2017humanpose}, allowing to focus the attention of the discriminator into the shape of the skeleton.
 
\vspace{1mm}
\noindent{\bf Motion discriminator}. the Base discriminator sees the sequences as absolute coordinates of joints in the space, and the EDM discriminator sees them as relative coordinates w.r.t the other joints. But these discriminators are missing the joint correlations between the absolute motion and their relative (articulated) counterpart. Thus, we consider a third discriminator that operates over the concatenation of both, the temporal differences of absolute coordinates $\|\bS^{out}(t) - \bS^{out}(t-1) \|_1$ and the temporal differences of EDM representations $\|\textrm{EDM}(\bS^{out}(t)) - \textrm{EDM}(\bS^{out}(t-1)) \|_1$, where $\bS^{out}(t)$ indicates generated the sequence at time $t$.

\subsection{Losses}
To train our network we use two main losses: 1) The reconstruction losses,  that encourage the generator to preserve the information from the  visible part of the sequence; 2) The GAN loss, which   guides the generator to inpaint the sequences by learning and reproducing the motion in the dataset. For all the following formulae, let $\bS$ be the input motion sequence, $\bM$ the occlusion mask, $\bS^{out}$ the generated sequence, $F$ number of frames and $J$ number of joints.

\vspace{1mm}
\noindent{\bf Reconstruction Loss}. 
Our default reconstruction loss computes the L2 norm over the generated sequence w.r.t. the visible portion of the ground truth.
\begin{equation}
\label{eq:rec_loss}
\mL_{rec} = \|(\bS^{out} \circ \bM) - (\bS \circ \bM)\|_2
\end{equation}
This loss is only applied over the visible part of the original and generated sequences. By doing this, we penalize deviations from the visible part of the sequences, while avoiding to penalize the different possible completions of the sequence.



\vspace{1mm}
\noindent{\bf Limb Distances Loss}.
~\cite{hernandez20173d} showed that most common actions can be recognized  from just the relative distance between the extremities, \ie  hands, feet and head. We therefore add a loss that explicitly enforces the correct distance between these semantically important joints. 
Since this loss looks at the relative distance, instead of the absolute position, it provides different gradients to the reconstruction loss, and encourages the network to learn a more precise location for the limbs. 
Formally, if we denote by $\mE=\{i,j\}$ the set of limb   pairs ,  the loss $\mL_{limb}$ is computed as:
\begin{equation}\label{eq:ext_loss}
 \sum_{f=1}^{F}  \sum_{\{i,j\} \in \mE}  \|\| \bS^m_{fi:} - \bS^m_{fj:}\|_2 - \|\bS^{m,out}_{fi:} - \bS^{m,out}_{fj:}\|_2\|_2
\end{equation}
where $\bS^{m,out}= \bS^{out}\circ \bM$, denoting again that this loss is only computed over the visible part of the original sequence.

\vspace{1mm}
\noindent{\bf Bone Length Loss}. We also enforce constant bone length of  the whole generated sequence. Its main goal is to  discourage the generator to explore solutions where the skeleton is not well formed. If we denote by $\overline\mB=\{\overline{l}_1,\ldots,\overline{l}_{B}\}$ the mean length of the $B$ body bones computed over the visible part of the sequence,  and by $\mB_f=\{l_{f1},\ldots,l_{f_{B}}\}$ the length of the bones at frame $f$, this loss is computed as
\begin{align}\label{eq:bone_loss}
\mL_{bone}=\sum_{f=1}^F\sum_{b=1}^{B}\|\overline{l}_{b}-l_{fb}  \|_2
\end{align}


\vspace{1mm}
\noindent{\bf Regularized Adversarial Loss}.
Our adversarial loss is based on the original GAN loss~\cite{goodfellow2014generative}, with the R1 regularization described in~\cite{mescheder2018training}. 
Let $G_\theta$ be the generator network, parameterized by the variable $\theta$, $D_\psi$ be the discriminator network, parameterized by the variable $\psi$, and $\mathbb{P}_o$ the distribution of input motion sequences.  We can then write the Discriminator Loss as:
\begin{align}\label{eq:disc_loss}
\mL_D =& \mathbb{E}_{\bS^{out} \sim \mathbb{P}_o}\left[\log(1 - D_\psi(G_\theta(\bS \circ \bM)))\right]\\
+ & \mathbb{E}_{\bS \sim \mathbb{P}_o}\left[\log(D_\psi(x))\right]
+ \frac{\gamma}{2} \mathbb{E}_{\bS \sim \mathbb{P}_o}\left[\|\nabla D_\psi(x)\|^2\right] \nonumber
\end{align}

The Generator Loss is as follows:
\begin{align}\label{eq:gen_loss}
\mL_G(\theta,\psi)= \mathbb{E}_{\bS^{out} \sim \mathbb{P}_o}\left[\log(D_\psi(G_\theta(\bS \circ \bM)))\right]
\end{align}

\begin{figure*}[t!]

\begin{tabular}{ccc}
\hspace{1.2cm}{\bf Frame Encoder} & \hspace{2.2cm}{\bf Generator U-Blocks} & \hspace{1.2cm}{\bf Residual CNN}
\end{tabular}\\
\begin{centering}
    \includegraphics[width=0.90\textwidth]{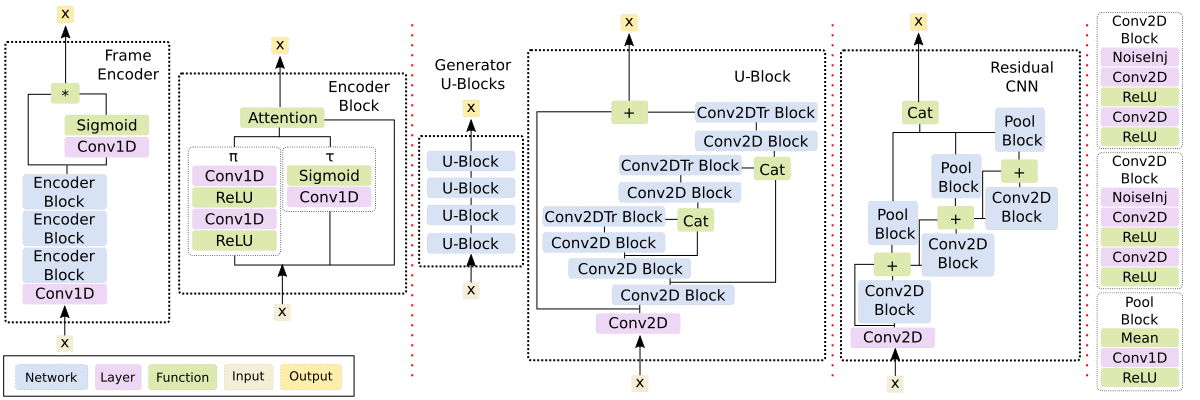}
    \caption{{\bf Details of the architecture.} From left to right: Frame Encoder; Generator U-Blocks; Residual CNN. In each case we plot the general view of the block (left) and the fine detail of the structural elements (right). The Attention function is defined as: $att(x, \pi, \tau)=\pi \tau + x(1-\tau)$. The Conv2DTr denotes Convolution 2D Transpose, also called deconvolution.
    }
    \label{fig:motiongan_details}
\end{centering}
\end{figure*}

\vspace{1mm}
\noindent{\bf Full Loss.}
The full loss $\mL$ consists of a  linear combination of all previous partial losses:
\begin{align}
\mL =& \lambda_r\mathcal{L}_{rec}+\lambda_l\mathcal{L}_{limb}+\lambda_b\mathcal{L}_{bone}+\lambda_D\mL_D+\lambda_G\mL_G
\end{align}
where $\lambda_r$, $\lambda_l$, $\lambda_b$, $\lambda_D$ and $\lambda_G$ are the hyper-parameters that  control the relative importance of every loss term. Finally, we can define the following minimax problem: 
\begin{equation}
G^\star =\arg \min_{G} \max_{D \in \mathcal{D}} \mathcal{L} \,\, ,
 \end{equation} 
where $G^\star$ draws samples from the data distribution.

\section{Metrics for motion prediction}
\label{sec:metric}



Our goal then is to analyze the distribution generated by our model, for which we propose metrics that are analogue to the Inception Score~\cite{salimans2016improved} and the Frechet Inception Distance~\cite{heusel2017gans}. With this in mind we propose the following metrics:

\vspace{1mm}
\noindent{\bf PSEnt}
measures the entropy in the power spectrum of a dataset. This metric can give us a rough estimate of the fitness of the model.
First we compute the power spectrum of the dataset independently per each joint and axis. Each joint-axis combination is considered a distinct feature of a sequence. Formally, the power spectrum of a feature is computed as: $PS(s_f) = \|FFT(s_f)\|^2$.
We can then compute the Power Spectrum Entropy over a dataset:
\begin{equation}
\label{eq:psent}
PSEnt(D) = \frac{1}{S}{\sum_{s\in D} \frac{1}{F}{\sum_{f=1}^{F} - \sum_{e=1}^{E} \|PS(s_f)\| * log(\|PS(s_f)\|)}}
\end{equation}
where $D$ is a dataset, $s$ is a sequence, $f$ is a feature, and $e$ is frequency.

A common characteristic of the generative models trained with the L2 loss is that they have the tendency to regress to the mean, lowering the entropy of the generated sequences. An entropy value lower than the expected is a telltale sign of a biased model, whereas a higher entropy value points to a rather noisy and maybe inaccurate or unstable model.

\vspace{1mm}
\noindent{\bf PSKL}
measures the distance (in terms of the KL divergence) between the ground truth and generated datasets:
\begin{equation}
\label{eq:pskl}
PSKL(C, D) =  \sum_{e=1}^{E} \|PS(C)\| * log(\frac{\|PS(C)\|}{\|PS(D)\|})
\end{equation}
where $C$ and $D$ are datasets, $s$ is a sequence, $f$ is a feature, and $e$ is frequency. The KL divergence is asymmetric, so we compute both directions PSKL(GT, Gen) and PSKL(Gen, GT) to have the complete picture of the divergence. If both directions are roughly equal, it would mean that the datasets are different but equally complex. On the other hand if the divergences are considerably different, it would mean that one of the datasets has a biased distribution.

\vspace{1mm}
\noindent{\bf L2 based metrics.} We also measure the distance between the ground truth sequence $s_{gt}$ and the generated  sequence $s_{gen}$, considering each joint ($j$) as an independent feature vector.
\begin{equation}
\label{eq:l2_metric}
L2(s_{gt}, s_{gen}) = \frac{1}{J}{\sum_{f=1}^{J} \|s_{gt} - s_{gen}\|_2}
\end{equation}
In~\cite{fragkiadaki2015recurrent, martinez2017human} $s$ is represented in Euler angles, but in our work $s$ is in coordinates, making it readable in millimeters. The mean is used to obtain a measure for the complete dataset.

\section{Implementation Details} 
\label{sec:implementation}
We next describe the blocks of our architecture. Code Available at: https://github.com/magnux/MotionGAN

\vspace{1mm}
\noindent{\bf Frame Autoencoder}.
The Frame Encoder (Fig.~\ref{fig:motiongan_details}-left) is a fully connected network with identical sequential blocks. Each block contains two consecutive fully connected layers, and an attention mechanism~\cite{barone2016low} at the end. The fully connected layers can also be seen as 1D convolutions with kernel size 1 along the time dimension. The attention is performed by applying a mask over the output of the block. The mask is a linear transformation of the input to the block, followed by a sigmoid activation. 
After the blocks, a final linear transformation and an attention are applied. The architecture is similar to a VAE~\cite{kingma2013auto}, but without the Gaussian constrains over the output of the encoder. The decoder network is a symmetrical network to the encoder, with the same number of blocks and layers, but independent parameters.

\vspace{1mm}
\noindent{\bf Generator U-Blocks}.
The goal of the generator is to produce an output that should be indistinguishable from an unmasked input, while preserving the shape of the sequence. To accomplish this, we use U-blocks~\cite{ronneberger2015u}  (see Fig.~\ref{fig:motiongan_details}-center) with convolutions that halve the spatial resolution of the input in each layer, until it reaches a small representation. Then a transposed convolution is used to double the resolution until it reaches again the same dimensions as the input. A key component in this architecture are the skip connections that connect the output of the convolutional layers to the input of the deconvolutional layers. We can think of this architecture as an iterative refinement, in which the output of a block is refined by the next block to produce a better final output. Following~\cite{karras2018style}, we also incorporate a noise injection layer into our convolutional blocks which makes the model prediction non-deterministic and enriches it.



\vspace{1mm}
\noindent{\bf Residual CNN}. 
We designed the architecture of our discriminator inspired on ResNet~\cite{he2016deep} and DenseNet~\cite{huang2017densely}.
Our network (Fig.~\ref{fig:motiongan})  branches in three discriminators, and each discriminator has a classifier, with the same architecture but   separate parameters. Their architecture  (Fig.~\ref{fig:motiongan_details}-right)  consists of  several consecutive blocks with two convolutional layers and additive residual connections, similar to ResNet. The outputs of each block are also transformed and then concatenated into a tensor. The final output is the concatenation of the outputs of all blocks. This output is finally passed onto a fully connected network that assigns a score. 


\vspace{1mm}
\noindent{\bf Spatial Alignment}. Since we are working over an absolute coordinate system, the sequences have a wide range of values (from mm to m). To improve the robustness of the generator we subtract the position of the hip joint in the first frame to all the joints in the sequence. Then the skeleton is rotated to always face in the same direction. The alignment is performed by a custom layer in the network before the frame encoder, and is reversed just after the frame decoder.

\section{Experiments}
\label{sec:experiments}

\begin{table}[t!]
\setlength{\tabcolsep}{3pt} 
\setlength\arrayrulewidth{0.9pt}
\centering
\resizebox{7cm}{!} {
	\begin{tabular}{| r| c | c | c |}
		\hline
		\multicolumn{1}{|r|}{\cellcolor[gray]{0.95}  Model       } & 
		\multicolumn{1}{c|}{\cellcolor[gray]{0.95}  PSEnt        } &
		\multicolumn{1}{c|}{\cellcolor[gray]{0.95}  PSKL(GT,Gen) } &
		\multicolumn{1}{c|}{\cellcolor[gray]{0.95}  PSKL(Gen,GT) } \\			
		\hline
		\multicolumn{4}{|c|}{0 to 1 second}\\
		\hline
		Org.Data. (Val vs Train)               & 0.67990 & 0.00590 & 0.00572         \\
		\hline
		Res.sup.~\cite{martinez2017human}        & 0.37492 & 0.03293 & 0.04524         \\
		\hline
		NoGAN                                  & 0.44363 & 0.03729 & 0.05040         \\
        Base disc                               & 0.73626 & 0.01198 & 0.01149         \\
        +EDM disc                              & 0.57045 & 0.01801 & 0.02131         \\
        +Motion disc                           & 0.72617 & 0.01220 & 0.01141         \\
        STMI-GAN                               & \bf0.68099 & \bf0.01090 & \bf0.01125         \\
        \hline
		\multicolumn{4}{|c|}{1 to 2 seconds}\\
		\hline
		Org.Data. (Val vs Train)               & 0.67749 & 0.00628 & 0.00611         \\
		\hline
		Res.sup.~\cite{martinez2017human}        & 0.20975 & 0.10188 & 0.17004         \\
		\hline
		NoGAN                                  & 0.27969 & 0.07989 & 0.13743         \\
        Base disc                               & 0.60450 & 0.01559 & 0.01766         \\
        +EDM disc                              & 0.49198 & 0.02546 & 0.03315         \\
        +Motion disc                           & 0.72963 & 0.01223 & 0.01129         \\
        STMI-GAN                               & \bf0.68328 & \bf0.01041 & \bf0.01010         \\
        \hline
		\multicolumn{4}{|c|}{2 to 3 seconds}\\
		\hline
		Org.Data. (Val vs Train)               & 0.67391 & 0.00640 & 0.00620         \\
		\hline
		Res.sup.~\cite{martinez2017human}        & 0.12752 & 0.17402 & 0.33566         \\
		\hline
		NoGAN                                  & 0.34717 & 0.06099 & 0.09562         \\
        Base disc                               & 0.60804 & 0.01396 & 0.01611         \\
        +EDM disc                              & 0.45627 & 0.03368 & 0.04596         \\
        +Motion disc                           & 0.72368 & 0.01312 & 0.01201         \\
        STMI-GAN                               & \bf0.71778 & \bf0.01306 & \bf0.01213         \\
        \hline
		\multicolumn{4}{|c|}{3 to 4 seconds}\\
		\hline
		Org.Data. (Val vs Train)               & 0.67891 & 0.00590 & 0.00566         \\
		\hline
		Res.sup.~\cite{martinez2017human}        & 0.09333 & 0.18692 & 0.37605         \\
		\hline
		NoGAN                                  & 0.26750 & 0.08672 & 0.15567         \\
        Base disc                               & 0.50224 & 0.02646 & 0.03460         \\
        +EDM disc                              & 0.41653 & 0.04516 & 0.06541         \\
        +Motion disc                           & 0.76111 & 0.01436 & 0.01275         \\
        STMI-GAN                               & \bf0.70985 & \bf0.01108 & \bf0.01024         \\ 
        \hline
		\multicolumn{4}{|c|}{0 to 4 seconds}\\
		\hline
		Org.Data. (Val vs Train)               & 1.65373 & 0.01225 & 0.01227         \\
		\hline
		Res.sup.~\cite{martinez2017human}        & 0.85732 & 0.13320 & 0.15644         \\
		\hline
		NoGAN                                  & 1.07468 & 0.10245 & 0.12508         \\
        Base disc                              & 1.58270 & 0.02197 & 0.02274         \\
        +EDM disc                              & 1.22901 & 0.08416 & 0.09894         \\
        +Motion disc                           & 1.77806 & 0.02434 & 0.02270         \\
        STMI-GAN                               & \bf1.69147 & \bf0.01888 & \bf0.01801         \\ 
		\hline
	\end{tabular}}
	\vspace{1.0mm}
	\caption{\textbf{Ablation Study.} Power Spectrum based metrics for different configurations of our model.}
	\vspace{-4.0mm}
	\label{tab:ablation}
\end{table}

\vspace{1mm}
\noindent{\bf Datasets}. In the experimental section we mainly use the Human3.6M~\cite{h36m_pami} dataset. We follow the same split used in~\cite{fragkiadaki2015recurrent, martinez2017human}.

\subsection{Motion Prediction}
In this section we compare our approach to~\cite{martinez2017human},  one of the baseline works in the state of the art. From this work, we are using the residual supervised model, which is a sequence-to-sequence model with residual connections and uses the labels as part of the inputs. Since our model is based on Cartesian coordinates, we compute the joint angle representation equivalent to that used by the Residual supervised (Res.sup.)~\cite{martinez2017human} model. This transformation allows us for a consistent comparison between the two.
	
We next run an ablation study, using always the same generator network, but training it with different discriminator networks. As we argued in previous sections, we aim to capture the distribution of the Ground Truth (GT) data, and each component in the architecture was designed with this purpose. Our hypothesis is that an adversarial loss is better than L2 and geometric losses to train a generative network. Even more, we argue that the complexity of the discriminator network should be correlated with a better result in the generated sequences. 

Our tested models are: NoGAN: generator network trained with the reconstruction losses over the whole sequence, and no adversarial loss. Base disc: is the same network, but using the encoder discriminator as loss for the generated part. +EDM disc: the network is trained using both the base and the EDM discriminators. +Motion disc: the network is trained using the base and motion discriminator. STMI-GAN: is the full network, trained with all joint discriminators.

Every discriminator seems to be adding information to the generated distribution. We can observe this qualitatively, and we can confirm it with the proposed metrics.

\noindent{\bf Entropy Analysis and KL distance Analysis}.
We should first note that the PSEnt in the original distribution is almost identical in every one second window, at approximately $0.678$. This number represents the entropy of the uniform distribution, which means that the short term frequencies are fairly uniform. The PSEnt raises to $~1.65$, when we consider a 4 second time window. Such raise means that the long term motion has a biased, and more complex frequency distribution, which is not uniform but denser in some parts of the spectrum. 

We can observe in Tab.\ref{tab:ablation} that the Res.sup.~\cite{martinez2017human} and NoGAN baselines decay in entropy as the seconds pass. Also we see that the KL divergence grows rapidly for the baselines and is around an order of magnitude higher than any of the GAN models. 
The GAN models all seem to have a good behavior, with PSEnt values close to the GT distribution. The Base disc model is already stable, but has some decay in entropy towards the end of the sequence. The Motion disc model seems to be pretty stable but it consistently overshoots on the entropy. This may be interpreted as the model overemphazising in moving. The EDM disc model seems to be harming at a first glance, since it considerably lowers the PSEnt, but the main point of adding this discriminator is to prevent unexpected poses from happening. It is a regularizer by design.

When we combine the three discriminators in the STMI-GAN model, the network approximates closely to the expected distribution. The STMI-GAN is stable both in PSEnt and PSKL and its performance does not decay as time passes. Indeed, it has a PSEnt close to the GT and a low PSKL, meaning that it is not only producing the same amount of motion but also the same kind of motion.
We should note that the Human3.6M dataset has a considerable difference between the validation and train splits when following the standard protocol~\cite{martinez2017human}. The PSKL between the validation and train splits for the whole sequence is around $0.012$, almost symmetrical, and the PSKL between the generated distribution of STMI-GAN and validation is around $0.018$, also symmetrical. This means that the distribution produced by the GAN is almost as close as the training and validation sets are.

\begin{figure}[t!]
\begin{centering}
    \includegraphics[width=0.40\textwidth]{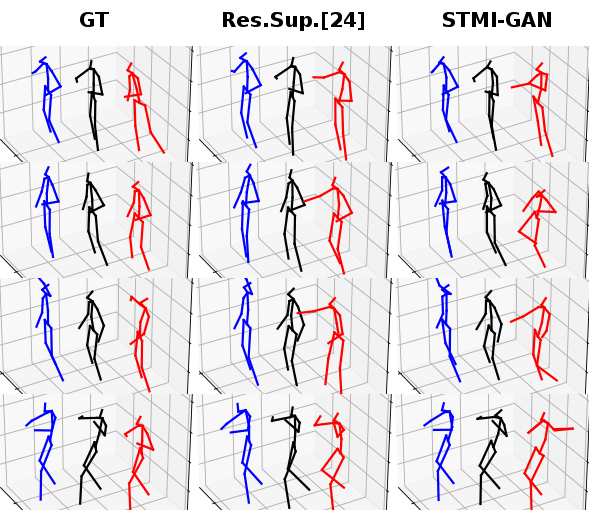}
    \caption{{\bf H3.6 Examples.} Blue is first frame, black is first predicted frame, red is last predicted frame. Total length is 4 secs, 2 seed + 2 predicted.}
    \vspace{-4.0mm}
    \label{fig:examples}
\end{centering}
\end{figure}

\noindent{\bf L2 metric experiments}. 
To demonstrate the point that L2 metric is not correlated with a realistic generation, we use a subset of the 120 test sequences of~\cite{fragkiadaki2015recurrent,martinez2017human}, concretely the  \#8, \#26, \#27, \#88. Fig.~\ref{fig:examples} shows the results of our approach and Res.sup.~\cite{martinez2017human} on these sequences. Note that   Res.sup.  has a tendency to converge to the same pose (see the red skeleton in the center column), which is  very close  to the mean pose in the dataset. There is also the tendency to produce very small motions (see black and red at the bottom center frame). Indeed, ~\cite{martinez2017human} shows that the zero velocity baseline is often better than their  model,  specially for the class 'discussion' which has high uncertainty (see last row in Fig.\ref{fig:examples}).

When  computing the L2 metric over angles as in~\cite{martinez2017human} we obtain the following results:  L2 Res.sup. $\rightarrow$ (0.69, 0.36, 0.64, 0.25);  L2 STMI-GAN $\rightarrow$ (1.09, 0.74, 1.33, 0.96). Despite the baseline model has a lower L2, the sequences generated by our approach seem more diverse and realistic. These effects derive from the objectives used to train each model. The baseline models aim to minimize the spatial distance through the  L2 loss. In our work we seek to reproduce the the distribution of human motion, and we use a GAN for this purpose. These objectives are not always aligned, and the L2 metric often fails to grasp the complexity of realistic human motions.

\noindent{\bf Noise Injection}.
It may seem that the noise injection would cause big differences in the output of the network, but actually the expected difference in the prediction is around $0.81mm$ per joint. This means that when called with the same seed sequence, the network produces almost identical sequences, only tweaking minor aspects of the sequence. This result confirms the effects of the noise reported in~\cite{karras2018style}. It is also interesting to note that the difference increases with the length of the prediction, meaning that indeed the injected noise is solving some level of uncertainty, but the maximum difference that we have measured predicting 4 seconds is $3.02mm$ per joint.

\begin{figure*}[t!]
    \includegraphics[width=1.0\textwidth]{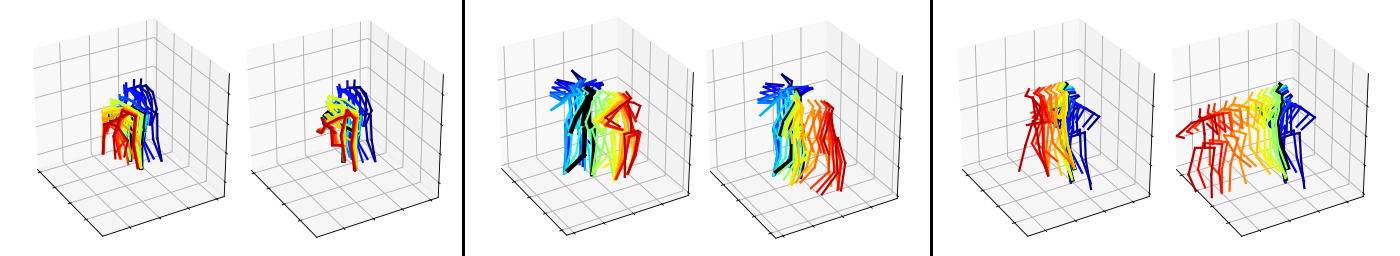}
    \caption{{\bf Example result.} Three examples  of the predicted motions (left: ground truth, right: predicted). The bluish colors are the part of the sequence that is observed. The prediction starts after the black skeleton and corresponds to the yellow-reddish colors.}
    \vspace{-4.0mm}
    \label{fig:qualitative}
\end{figure*}

\vspace{1mm}
\noindent{\bf Qualitative Evaluation}.
We conducted evaluation with 15 person and four distinct surveys, all of them following the same scheme: a prediction model vs the ground truth. In the first two surveys we tested the baseline Res.sup. and  STMI-GAN, to perform relative motion prediction. In the case of our model we removed the translation from the prediction to make it comparable. The last two surveys assess the absolute motion prediction. We use our NoGAN model as baseline vs our STMI-GAN model. 
In these surveys our goal is to obtain a 50\% chance of being classified as real. More than 50\% would mean that the model is "more realistic" than the ground truth.
As we can observe in Tab.~\ref{tab:survey} the results have a wide range of values. This is due to the fact that the survey was sent to a diverse audience. 
We can see that the baselines perform in a similar range of average values as our model, our model being bit better. However, the baselines perform very poorly to the trained eye, as the min score tells us.
It is worth to note that while the relative motion prediction is an easier problem compared to the absolute motion prediction, the average score of our STMI-GAN is lower in the relative setting. This suggests that the relative motion generation is both harder for machine learning models and for humans. Also, the highest score on the surveys is in the max of the STMI-GAN in absolute prediction. This implies that some individuals were very convinced by the results of our model.

  \begin{table}[t!]
	\setlength{\tabcolsep}{3pt} 
	\setlength\arrayrulewidth{0.9pt}
	\centering
	\resizebox{6.5cm}{!} {
		\begin{tabular}{|r |c|c|c|c|}
			\hline
  \multicolumn{1}{|c|}{\cellcolor[gray]{0.95} Model } & \multicolumn{1}{c|}{\cellcolor[gray]{0.95} Motion}&
  \multicolumn{1}{c|}{\cellcolor[gray]{0.95} Min} & \multicolumn{1}{c|}{\cellcolor[gray]{0.95} Avg} &
  \multicolumn{1}{c|}{\cellcolor[gray]{0.95} Max} \\
 			\hline
 			\hline
Res.sup~\cite{martinez2017human}     & Rel.   & 6.25\%  & 31.88\%          & 40.63\% \\
STMI-GAN                             & Rel.   & 25.00\% & 33.54\%          & 40.63\% \\
NoGAN                                & Abs.   & 9.38\%  & 31.46\%          & 62.50\% \\
STMI-GAN                             & Abs.   & 15.63\% & \textbf{38.39}\% & 62.50\% \\
        \hline
  		\end{tabular}}
		\vspace{1.0mm}
		\caption{\textbf{Human Evaluation.} Percentage of times that a human evaluator thought a generated sequence was real. The min is the score of the "hardest" evaluator, who was fooled the less by the generative models. The max is the score of the "easiest" evaluator, who was confused more often by the model.}    
		\vspace{-4.0mm}
		\label{tab:survey}
	\end{table}

Fig.\ref{fig:qualitative}   shows three examples. The sequence on the left is easy to predict, just continuing the motion of picking up things off the floor. The sequence on the center is a bit harder, as it is easy to guess that the person will continue walking, but the person halts after a couple of steps, and this is unexpected. The sequence on the right is challenging, because just before the generator begins its forecast, the person stops. This makes it hard to predict as the uncertainty rises and many options are plausible. We can see that the generator in fact guessed the action (walking) and the correct direction, but the speed of the motion is not accurate. We consider it however a good guess given the input.

 \subsection{Occlusion Completion}
Finally, in  Table~\ref{tab:occlusion} we report the robustness to different types of occlusion. In every test we used 80\% of occlusion, \ie  we are trying to recover a sequence conditioning only on  20\% of the data. The generator model is particularly robust to structured occlusions, it produces good results even without the GAN, we hypothesize that this is because it was trained to produce anthropomorphic guesses. When the occlusions happen at random, linear interpolation is also a good approach, but depending on the nature of the occlusion we may need a more robust model.

\section{Conclusions}
\label{sec:conclusions}

We have presented a novel GAN architecture  to  predict 3D human motion from  historical 3D skeleton poses. We have extended existing works, by also forecasting (beyond 2 seconds) the absolute position of the body. We have formulated our problem as an inpainting task  in   spatio-temporal volumes. In order to capture the essence and semantics of the human motion, the training of our network has been mostly guided by three independent discriminators. They encourage the generation of motion sequences with a similar frequency distribution to that of the original dataset. Since L2 is known not to be adequate to compare generated sequences, we have also proposed new metrics that estimate the frequency distribution of the datasets, grasping the concept of multiple possible futures. Experimental results on Human3.6M show the effectiveness of our model to generate highly realistic human motion predictions. 

  \begin{table}[t!]
	\setlength{\tabcolsep}{3pt} 
	\setlength\arrayrulewidth{0.9pt}
	\centering
	\resizebox{8cm}{!} {
		\begin{tabular}{|r |c|c|c|c|}
			\hline
  \multicolumn{1}{|c|}{\cellcolor[gray]{0.95} Problem } &
  \multicolumn{1}{c|}{\cellcolor[gray]{0.95} Linear Int}&
  \multicolumn{1}{c|}{\cellcolor[gray]{0.95} LR-Kalman\cite{burke2016estimating}} &
  \multicolumn{1}{c|}{\cellcolor[gray]{0.95} NoGAN} &
  \multicolumn{1}{c|}{\cellcolor[gray]{0.95} STMI-GAN} \\
 			\hline\hline
Joint Occl.     & 232.06    & 329.23  & \bf96.52  & 108.99 \\
Limb Occl.      & 209.45    & 312.40  & \bf123.07 & 189.09 \\
Missing Frames  & \bf50.42  & 123.05  & 72.67     & 102.03 \\
Noisy Transm.   & \bf94.54  & 308.98  & 98.53     & 110.29 \\


        \hline
  		\end{tabular}}
		\vspace{1.0mm}
		\caption{\textbf{Occlusion Completion.} We test different types of occlusion, concretely: \textbf{Joint Occlusions:} joints occluded at random in each frame. \textbf{Limb Occlusions:} joint chains representing limbs are occluded at random in each frame. \textbf{Missing Frames:} entire frames are occluded at random. \textbf{Missing transmission:} data points in any dimension are occluded at random. The table reports the L2 metric over coordinates(See \ref{sec:metric}). }     
		\vspace{-4.0mm}
		\label{tab:occlusion}
	\end{table}

\vspace{1mm}
\noindent{\bf Acknowledgments:} 
This work is supported in part by an Amazon Research Award and by the Spanish MiNeCo under projects HuMoUR TIN2017-90086-R and Mar\'ia de Maeztu Seal of Excellence MDM-2016-0656.  Also was funded by the Deutsche Forschungsgemeinschaft  GA 1927/4-1 and the ERC Starting Grant ARCA (677650). 

{\small
\bibliographystyle{ieee_fullname}
\bibliography{references}
}

\end{document}